\documentclass[10pt,journal,compsoc]{IEEEtran}
\usepackage{soul}
\usepackage{ragged2e}
\usepackage{scrextend}

\ifCLASSOPTIONcompsoc
  \usepackage[nocompress]{cite}
\else
  \usepackage{cite}
\fi
\ifCLASSINFOpdf
   \usepackage[pdftex]{gra phicx}
\else
   \usepackage[dvips]{graphicx}
\fi
\usepackage{amsmath,amssymb,amsfonts}
\usepackage{algorithmic}
\usepackage{url}
\usepackage{hyperref}
\urldef{\urlA}\url{https://github.com/MJafarMashhadi/pprz_tester} 
\urldef{\urlB}\url{https://github.com/MJafarMashhadi/pprz_tester/wiki/Paparazzi-Patches}

\urldef{\urlC}\url{https://github.com/MJafarMashhadi/pprz_tester/tree/master/pprz_tester/pprzlink_enhancements}

\usepackage{textcomp}
\usepackage{dsfont}
\usepackage{tabularx}
\usepackage{multirow}
\usepackage{booktabs}
\usepackage{tcolorbox}
\usepackage{listings}
\usepackage{lscape}
\usepackage{subcaption}

\newtcolorbox{rqanswer}{%
    colback=gray!30,
    colframe=gray!45,
}

\begin{document}
%
\title{Deep State Inference: Toward Behavioral Model Inference of Black-box Software Systems}
%
%
%
%

\author{Foozhan~Ataiefard,
        Mohammad~Jafar~Mashhadi,
        Hadi~Hemmati
        and~Neil~Walkinshaw
\IEEEcompsocitemizethanks{\IEEEcompsocthanksitem F. Ataiefard, M.J. Mashhadi, and H. Hemmati, University of Calgary, Calgary,
AB, Canada.\hfil\break E-mails: foozhan.ataiefard1@ucalgary.ca, mohammadjafar.mashha@ucalgary.ca, hadi.hemmati@ucalgary.ca
\IEEEcompsocthanksitem N. Walkinshaw, University of Sheffield, Sheffield, UK.
E-mail: n.walkinshaw@sheffield.ac.uk}

\thanks{Manuscript received ??? ??, 2020; revised ??? ??, ????.}}

%
%

\markboth{IEEE Transactions on Software Engineering,~Vol.~, No.~, ~202x}%
{Mashhadi \MakeLowercase{\textit{et al.}}: Application of Transfer Learning to Lower The Cost of Black-box\\Behavioral Model Inference using Deep Neural Networks}
%



\IEEEtitleabstractindextext{%
\begin{abstract}
\justify
Many software engineering tasks, such as testing, debugging, and anomaly detection can benefit from the ability to infer a behavioral model of the software. Most existing inference approaches assume access to code to collect execution sequences. In this paper, we investigate a black-box scenario, where the system under analysis cannot be instrumented, in this fashion. This scenario is particularly prevalent with control systems’ log analysis in the form of continuous signals. In this situation, an execution trace amounts to a multivariate time-series of input and output signals, where different states of the system correspond to different ’phases’ in the time-series. From an inference perspective, the challenge is to detect when these phase changes take place. Unfortunately, most existing solutions are either univariate, make assumptions on the data distribution, or have limited learning power. Therefore, we propose a hybrid deep neural network that accepts as input a multivariate time series and applies a set of convolutional and recurrent layers to learn the non-linear correlations between signals and the patterns over time. We show how this approach can be used to accurately detect state changes, and how the inferred machine learning models can be successfully applied to transfer-learning scenarios, to accurately process traces from different products with similar execution characteristics. Our experimental results on two UAV autopilot case studies (one industrial and one open-source) indicate that our approach is highly accurate (over 90\% F1 score for state classification) and significantly improves baselines (by up to 102\% for change point detection). Using transfer learning we also show that up to 90\% of the maximum achievable F1 scores in the open-source case study can be achieved by reusing the trained models from the industrial case and only fine tuning them using as low as 5 labeled samples, which reduces the manual labeling effort by 98\%.

\end{abstract}

}

\maketitle

\IEEEdisplaynontitleabstractindextext

%
\IEEEpeerreviewmaketitle

\IEEEraisesectionheading{\section{Introduction}\label{sec:intro}}


\IEEEPARstart{A}{pproaches} to reverse-engineer and reason about software behaviour are often based upon dynamic analysis methods. These involve the collection of execution traces\cite{ammons2002mining} - sequences of execution events or data from which the model can be inferred. Collecting these traces either requires some form of source code instrumentation (i.e. the systematic insertion of logging statements), or relies upon features of the execution environment (e.g. Reflection in Java).  These methods are especially helpful for unit-level analysis, particularly when source code is available and the performance overheads incurred by tracing are negligible.

Such tracing approaches can however be impossible (or at best highly impractical) to apply in larger systems, especially if they incorporate black-box components. Black-box components can be impossible to instrument and inspect, and might even be physically sealed (such as embedded components). Aside from the challenges of observability, large-scale systems can be prohibitively expensive to trace \cite{mashhadi2019empirical}, and tracing can incur performance overheads that lead to observed behaviour that deviates from the non-traced equivalent - an issue that is especially problematic for real-time systems \cite{musuvathi2008finding}. 

Instead, the task of tracing such large systems is often limited to `lightweight' alternatives; passively recording the observable state-variables of the system without accessing any implementation details. This form of analysis is particularly useful for control systems, which tend to comprise large numbers of black-box components. In a car, for example, we might not have access to the internals of its cruise control system, but we \emph{can} readily monitor state variables such as the extent to which the throttle or break is applied, and its speed and acceleration. Traces that are collected in this way tend to take the form of multivariate time-series, where each state variable corresponds to a signal.

One key difference from `traditional' software traces is that these more continuous traces are not associated with internal discrete events, which means that there is no clear indicator of (1) what the main states of the system are, and (2) when state changes occur. In our cruise control example, the trace of the system does not include internal calls to indicate when the cruise control is accelerating the car or decelerating it, or when it deactivates itself. This has to be discerned from the multivariate signals in the trace.

In this paper we present a technique that can address these questions. Our solution involves training a hybrid deep learning model (including convolution and recurrent layers) on the time-series to predict the state of the system at each point in time. The deep learning model automatically performs feature extraction, which makes it much more effective and flexible than traditional methods. In addition, we do not make any assumption about statistical properties of the data which makes it applicable to a wide range of subjects.

We applied and evaluated this method on an autopilot software used in an Unmanned Aerial Vehicle (UAV) system developed by our industrial partner, Winnipeg-based MicroPilot Inc. We then replicated the results on another highly capable and widely used autopilot, Paparazzi \cite{hattenberger2014using}.
We evaluated the method from two perspectives: (1) how well the model can detect the point in time at which a state change happens, and (2) how accurately it can predict which state the system is in. 
We also experimented with non-hybrid architectures to see how much does the hybrid architecture contribute to the overall performance, and explored role of hyper-parameter tuning on performance. 
Finally, we also explored the application of transfer learning to lower the cost of data labeling which is the most expensive step in this approach.

Our results indicate that the approach outperforms the state of the art approaches, both in terms of state-change (or `change-point') and state detection. In the MicroPilot case study we observed improvements ranging between 88.00\% and 102.20\%
 in the F1 score compared to traditional change-point detection techniques. For the state classification we saw improvements in the range from 7.35\% to 16.83\% in the F1 score compared to traditional sliding-window classification algorithms. 
For the Paparazzi case study we witnessed smaller (albeit still substantive) improvements for the change-point detection accuracy in the range between 13\% and 43\%, and a much larger improvement for the state-detection accuracy in the range from 77.20\% to 87.97\%.

We also observed a significant reduction of manual labeling cost, when using transfer learning, which achieves up to 90\% of the potential F1 scores by only 2\% of dataset being labeled (only 5 test cases). 

The contributions of this paper can be summarised as follows:
\begin{itemize}
    \item A deep learning architecture to to identify states and state changes of the black-box software systems under study towards behavioural model inference.
    \item An empirical evaluation that demonstrates the accuracy of our approach with respect to two real-world and large-scale case studies, involving a UAV autopilot system developed by our industry partner as well as an open source UAV autopilot.
    \item A hyper-parameter tuning pipeline to optimize the performance of the deep learning model.
    \item A transfer learning approach to reuse the pretrained models, as much as possible, and reduce the manual labeling cost.
\end{itemize}
We have also made An automated fuzz testing tool capable of generating and executing test cases for Paparazzi autopilot. All source codes, models, execution scripts, and all of the unrestricted data are available online\footnote{\label{foot:hybrid}\url{https://github.com/sea-lab/hybrid-net}, Due to confidentiality, the MicroPilot dataset cannot be shared publicly}.
The rest of this paper is organized as follows: 
In section~\ref{sec:background} we go through some background material and related papers this work is based on. 
In section~\ref{sec:approach}, we present our proposed method and model. The way it was evaluated and the results are presented in section~\ref{sec:experiment}. 
We provide a summary of the paper and briefly discuss some paths for future continuation of this work in section~\ref{sec:summary}.

\section{Background and Related Work} \label{sec:background}

We start this section with a description of the analysis scenario that motivates our work. We then cover the related work in the dynamic analysis of software behaviour and in time-series analysis. This is followed by an overview of the Deep Learning and transfer learning techniques that we will be drawing upon in our own solution.

\subsection{Motivating scenario}
\label{sec:motivation}
In this paper we consider the task of carrying out a system-level analysis of a Cyber-Physical System (CPS). By this we refer to a system, where the high level behaviour is controlled by a network of hardware and software components. The behaviour of the system is also time-sensitive; a small delay in the execution of one component can have a significant impact. 

To analyse these systems, we cannot employ conventional tracing approaches that involve instrumentation and logging, because this may be physically impossible, and could incur an unacceptable performance overhead \cite{mashhadi2019empirical,musuvathi2008finding}. Instead, we merely rely on the ability to record the data that is available anyway - the external input and output data-streams, perhaps coupled with hardware registers that passively provide information about the system (e.g. integrated circuits commonly include pins that automatically provide profiling information such as cache-misses or energy usage \cite{eljuse2018comparison}).

Our setting is motivated by a partnership with an autopilot manufacturer for UAVs to study their autopilot software. The goal was to determine its internal state from its input/output signals, over time. In this scenario, the inputs are the sensor readings going into autopilot and the outputs are command signals sent to controller motors of the aircraft, reflecting the reaction of the autopilot to each input at each state. A `state' in this example is the high-level stage of a flight (e.g. ``hold altitude'', ``landing'', ``climbing'', etc.), and a state change happens when the  input values trigger a constraint in the implementation (which is hidden) that changes the way the output signals are generated. Note that not all states are mapped one-to-one to a high-level intuitive domain concept. For example, there might be multiple states during a ``take-off'' scenario, which a non-expert can not give them a separate name. Therefore, in our study (similar to what is present in the source code) all states are referred to by state IDs, rather than a formal name.

During a flight, the autopilot monitors changes in the input signals and makes adjustments to its outputs in order to hold some invariants (predefined rules). 
For example, if it is in the ``hold altitude'' mode, it monitors the altimeter readings and makes proportionate adjustments to the throttle or the nose pitch to return to the desired altitude when it goes out of an acceptable range. In this respect it amounts to a typical feedback loop controller, such as a Proportional Integral Derivative controller (PID)  \cite{feedbacksystemsBook}.
When the state is changed (e.g. by a pre-loaded flight plan) from ``hold altitude'' to ``descend to X ft'', the set of invariants that guide the autopilot are changed. It means its reactions to variations in inputs will change. In this example, a decreasing altimeter reading will not trigger an increase in the throttle anymore. Therefore, it is often possible for a domain expert to visually determine what the state of autopilot is at each point in time, by monitoring input and output signals together.

\begin{figure}
\centering
  \begin{subfigure}{0.83\columnwidth}
  \centering
    \includegraphics[width=\columnwidth]{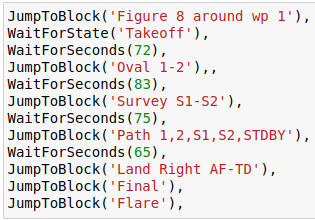}
    \caption{A sample test case (flight scenario) for UAV autopilot consisting of a series of commands for the autopilot to perform, which is executed in Paparazzi's flight simulator.} \label{fig:statesa}
  \end{subfigure}%
  \newline   
  \begin{subfigure}{\columnwidth}
  \centering
    \includegraphics[width=\columnwidth]{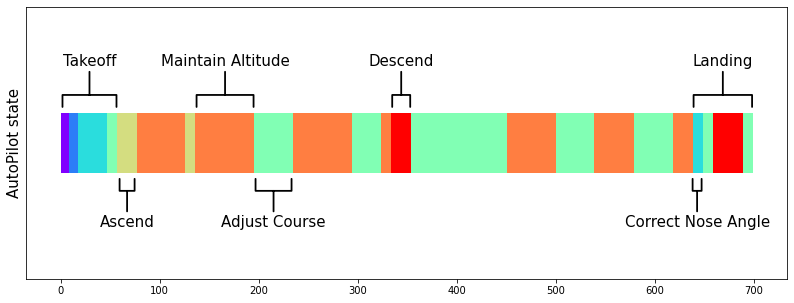}
    \caption{Visualization of different states in the above flight scenario. Each color represents a state ID that is assigned to an internal state. The state names are not officially provided and are our best guesses, for the sake of this example. The X-axis is the time axis (time steps).}
    \label{fig:statesb}
  \end{subfigure}%

\caption{An example flight scenario and its state change visualization.}
\label{fig:states}
\end{figure}

A sample test case for a flight scenario is shown in Figure~\ref{fig:statesa}. We have run this plan using the Paparazzi simulator and annotated the resulting flight scenario by observing plane's path and behaviour. Note that there are no formal state names in the source code. Some flight-states might not even translate to an intuitive domain concept or label. In this paper we use the simple visualization format shown in Figure~\ref{fig:statesb}, where different states simply correspond to different colors to provide a visual intuition of our prediction results.  

Although visualization is helpful to an extent, for applications that involve potentially large amounts of data (e.g. as a result of testing on a simulator), relying on a domain expert can become prohibitively expensive. Our goal is to automate this task. We assume that we have access to a labelled dataset (a manageable amount of data labelled by an expert, as was the case with our industrial partners), where the time-stamps signifying a state-change have been labelled in the trace. The challenge is to use this data to predict the corresponding states and state-changes in unlabelled data, collected from new additional executions of the system.

    
\subsection{Inference of State-Based Models}

For systems that exhibit sequential patterns of behaviour, such as the UAVs considered in this paper, state-based models are commonly used to capture all of the potential sequences in a succinct graphical representation. These can offer a powerful basis for verifying and validating behavioural properties.  Numerous reverse-engineering approaches have emerged that seek to infer models from samples of execution traces, starting with Biermann and Feldman's $k$-tails algorithm \cite{biermann1972synthesis} almost 50 years ago. 

Until relatively recently, the vast majority of these reverse-engineering techniques assumed that the traces are encoded as sequences of discrete labels. For example, labels for a UAV might include `take-off' or `land'. More recent approaches  have also enabled these events to be accompanied by corresponding state data values \cite{lorenzoli2008automatic, walkinshaw2016inferring, cassel2016active}.

In our setting, the trace is essentially a continuous multivariate time-series. There are no discrete markers to identify when a given state is entered and exited. Whereas our immediate goal is to convert these multivariate time-series into sequences of labels, this raises the prospect of opening up this family of systems to this well-established and diverse family of state-based model reverse-engineering techniques.

\subsection{Time-Series Analysis}

When an execution trace is encoded as a time-series, it can be analysed with a variety of signal-processing algorithms. This was first established (in a software engineering context) by Kuhn and Greevy \cite{kuhn2006exploiting}. In their case, a trace was encoded as a series of call-stack depths, producing a single series per execution. They then showed how signal-processing techniques could be used to cluster and compare different traces.

\subsubsection{Change Point Detection}

In our context we are especially interested in inferring which portions of a trace correspond to states, and which points in a trace correspond to a state-change. This latter task is referred to as `Change Point Detection' (CPD) \cite{aminikhanghahi2017survey}. It is a well-studied subject due to its wide range of applications \cite{basseville1993detection}.

In general, CPD algorithms consist of two major components: a) the search method and b) the cost function \cite{Truong2018ChangePointSurvey}.
Search methods are either exact or approximate. For instance, Pelt is the most efficient exact search method in the CPD literature, which uses pruning \cite{killick2012optimal}. Approximate methods include window-based \cite{basseville1993detection}, bottom-up \cite{keogh2001online}, and binary segmentation \cite{scott1974cluster}. 

The cost function measures the `goodness of fit' of a specific area of the signal to a given model. These can vary from simply subtracting each point from the mean to  more complex metrics, such as auto-regressive \cite{angelosante2012group} or kernel-based cost functions. Amongst kernel-based functions, linear and Gaussian kernels are the most popular \cite{Truong2018ChangePointSurvey}.

A multitude of techniques have been developed to tackle specific variations of the CPD problem \cite{oh2002analyzing,  reeves2007review}.
Some assume that time series has only one input variable (is univariate) \cite{fryzlewicz2014wild}, that is only a single change-point \cite{bai1998testing}, that the number of change points is known beforehand \cite{lavielle2005using}, or that the data obeys some specific statistical properties \cite{chen2011parametric, takeuchi2006unifying}. 

Unfortunately these assumptions are not (generally) applicable in our scenario. For our purposes, a CPD method should work on multivariate data, and be able to capture non-linear relations between signals without making restrictive presumptions about the underlying data. 
It also needs to be resilient to time lags between an input signal change and its effect on the output signal (and the system-state). 

\subsection{Deep Learning and its Application to Time Series Data}

Our research goal is to develop a technique that overcomes the current limitations of CPD techniques. It should accurately detect an arbitrary number of changes in multivariate time-series, without imposing limiting assumptions about statistical properties of the underlying signals. There have already been several successful examples of the application of Deep Learning techniques to time-series data, albeit not CPD -- this will be the subject of the technique that we present in Section \ref{sec:approach}. 

\subsubsection{Hybrid Deep Neural Networks}

One fundamental Machine Learning task is to identify salient features in data (e.g. to identify the properties within a multivariate time-series that indicate a change point). For non-trivial data-sets, these feature-encodings may not be readily available. They can also be difficult to determine from a human standpoint because they may arise from complex non-linear interactions within the data. The task of identifying these features automatically is referred to as \emph{feature extraction} \cite{murphy2012machine}. 

Convolutional neural networks (CNNs) present a particularly effective solution to this problem for sequential signals. These are neural networks where the hidden units are associated with `receptive fields' -- matrices that capture the local contextual data for a given area of the input. These are then processed by filters which are convolved with particular filters (also referred to as kernels) that amplify particular aspects of the data \cite{morales2016deep, zeng2014convolutional, yang2015deep}. The weights for hidden nodes are tied across the whole time-series. This means that useful features that are discovered in one zone can be re-used elsewhere without having to be independently learned \cite{murphy2012machine,lecun2015deep}. CNNs also exhibit `translation invariance', which means that they are able to classify patterns no matter where they occur in the input data (in our case, the time-series).

Recurrent neural networks (RNN) have shown great performance in analysing sequential data such as machine translation, time-series prediction, and time-series classification \cite{cho2014learning,wang2017time,Ordonez2016} 
RNNs can capture long-term temporal dependencies, a property that is especially useful in our case \cite{Che2018}. For example, they might learn that ``climb'' state in a UAV autopilot usually follows ``take off''. Therefore, while it is outputting ``take off'' it anticipates what the next state will probably be and as soon as its input features start shifting, it detects the onset of a state change. This has the potential to improve the predictive capabilities of the model and make it more efficient at detecting in a way that could be difficult to match with classic methods.
The combination of RNN with CNN is generally referred to as a `hybrid deep neural network' \cite{wang2017time}.

\subsubsection{Application of Deep Learning to Time Series Data} \label{sec:related_work_har}

One major driver in Time-series analysis has been the growth in devices such as fitness trackers and smart watches, which use sensors to collect data about an individual's movements or health-indicators such as heart rate. One challenge is to detect high-level episodes of distinctive human behaviour - e.g. to determine whether someone is running or walking from the accelerometer data in their phone. This particular challenge is referred to as Human Activity Recognition (HAR). Several attempts to apply deep learning to HAR tasks have shown that hybrid neural networks (the combination of CNN and RNN) are particularly effective \cite{deepsense,morales2016deep, Ordonez2016, zheng2016exploiting}

In the context of computer vision, there have been several approaches to automatically segment images into distinctive areas (e.g. to detect cells in a petri-dish \cite{ronneberger2015u}, or to detect distinctive phases of memory usage in a bitmap representation of sequential memory-bus accesses \cite{huffmire2006wavelet}). In this context, U-net has emerged as one of the promising auto-encoder architectures \cite{ronneberger2015u}.

Although U-net is specific to bitmap data, its principles have also been adapted for time-series analysis by Perslev et al., called U-time \cite{perslev2019u}. Their model is fully convolutional; so although it is good at recognising repeated localised patterns, the lack of recurrent cells means that it lacks the benefit of RNNs, particularly the ability to process long-term dependencies that may arise in a time-series.

\subsubsection{Transfer Learning}
\label{subsec:transfer_learning}
It is often the case that, as part of a software analysis process, the analyst has access to a separate piece of software with a similar functionality. It might be an older (more robust) version of the software, a similar product in the company's product line, an open source equivalent, or a similar product by a competitor. These alternative versions can be used for example in testing (for regression testing, or to address the test-oracle problem \cite{barr2014oracle}).

In these settings, Machine Learners can also benefit. `Transfer learning' \cite{pan2009survey} refers to a set of techniques to use the information that a machine learning model learned in a task, to improve its performance with respect to a different (but similar) task. Using transfer learning can result in a faster training time (reaching the asymptotic performance in fewer training epochs), a better result overall (a higher asymptote), requiring fewer training examples to reach an acceptable performance, a combination of these results and more. 

To understand transfer learning, in more details, let's look at a typical example from machine learning context. In the next section, we then show how this will apply in our software engineering problem in hand. Assume we have trained a DNN model, with very high accuracy, for an image classification task, on a large dataset of cats and dogs images. Let's call this the ``source'' problem. Now assume we have a ``target'' problem that is image classification for boats and planes. Here we have a small dataset of images that includes boats and planes and the task is to classify them. Obviously, with the small target dataset a DNN won't be effective. Also, we can not reuse the trained model from the source problem directly on the target problem, since the objective is different. To address this problem, we can use transfer learning to reuse the high accuracy model for classifying cats and dogs, as much as possible, on the new but similar problem of classifying boats and planes. 

The process that is also called fine tuning, is to first train the source model on the source dataset (image classification model on the cats and dog dataset, in the above example). Then freeze all layers of the trained model, except for the last few layers. By ``freezing'' a layer, we mean that we keep the trained coefficients constant throughout the subsequent training (fine tuning) steps. The last couple of layers have been chosen for fine tuning (being trained on the new dataset) since these may need to be adapted to the new task (e.g., the task of classifying new objects in the above example would require a different output layer from the original classification model). In this setting the final training model is reused as much as possible, hopefully retaining the properties that lead to its high accuracy in the source dataset, and leading to an improved performance for the new classification task. 

\section{Hybrid Neural Network for State Inference} \label{sec:approach}

 Although the analysis of time-series is well-established, Section \ref{sec:background} also shows that current approaches to the sequential analysis of time-series have limitations. In this section we present a technique that seeks to overcome these limitations by applying a Hybrid Neural Network. In this section we show how our technique can be used to infer the states of a software system from run-time data, recorded in the form of a time-series that is comprised of the values of inputs and outputs of the system. In the rest of this section we present the architecture of our approach, show how to encode data for it, and discuss its implementation.

\subsection{The Model Architecture}
\begin{figure*}
    \centering
    \includegraphics[width=\textwidth]{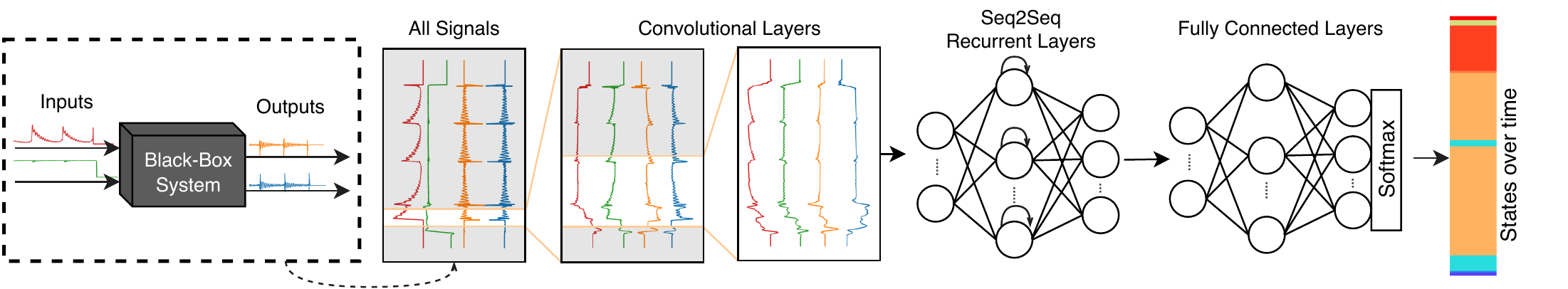}
    \caption{The input and output signals of the black-box system are captured as a multivariate time series; they are processed in a deep neural network that consists of 3 sections: convolutional, recurrent, and dense (fully connected) to predict the system's internal state and its changes over time.}
    \label{fig:general_net}
\end{figure*}

As can be seen in Figure~\ref{fig:general_net}, we capture inputs and outputs as a multivariate time series. To achieve its goal of accurately mapping phases in the time-series to high-level states, our solution needs to (1) automatically identify features within the data that indicate a state-change, and (2) accurately use these features to classify change-points and to identify which intervals between change points correspond to equivalent states in the data.

The architecture of our proposed model is illustrated in Figure ~\ref{fig:general_net}. The Convolutional Layers are intended to discover local features, such as sudden changes in phasal behaviour. Recurrent layers are used to process the sequential aspect of the data - to learn which specific states tend to occur in sequence. Finally, we use dense (fully-connected) layers to reduce the dimensions of the preceding layers to match the output dimensions. If there are only two states, the last layer can have a sigmoid activation function and be of shape $L$ (the length of the input), otherwise, to match the one-hot encoding of labels, an output of shape $L\times N_s$ with softmax activation along the second axis ($N_s$) is required ($N_s$ being the number of possible states).
In terms of loss function, we apply the dice overlap loss function, which is typically used in image semantic segmentation tasks \cite{milletari2016v, sudre2017generalised}. An important property of this loss function for our case is that it accommodates class imbalances.

\subsubsection{Configuration}

The number of different types of layers, filters and kernel sizes are hyper-parameters that should be selected based on the size of data and the complexity of the system under analysis. Using a sequence of convolutions with (a) increasing numbers of filters and the same kernel size, (b) the same number of filters and increasing kernel size, and (c)  decreasing numbers of filters increasing kernel sizes are all strategies that have been used by well-known architectures such as VGG and U-net \cite{simonyan2014very, ronneberger2015u}. Our model is configured as follows.

\begin{figure*}
    \centering
    \includegraphics[width=\textwidth]{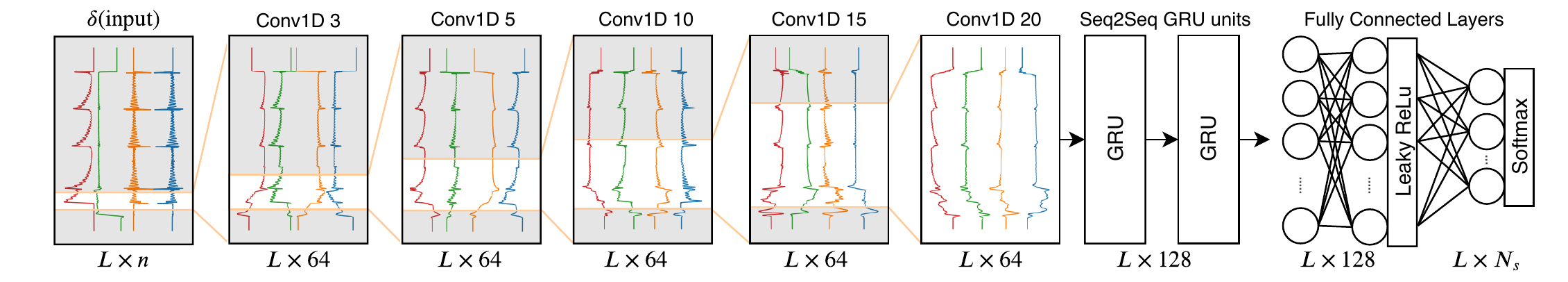}
    \caption{Model architecture in a nutshell. Tandem convolutional layers with increasing kernel size each with 64 filters) fed into two sequence-to-sequence GRU layers each with cell size of 128}, which is then fed into fully-connected layers to output the predicted system state, as a list of one-hot encoded states. 
    $L=18,000,\: N_s=25$.
    \label{fig:model_arch}
\end{figure*}

The first few layers of the model are convolutional layers. We used 5 convolutional layers with 64 filters each and a growing kernel size.
The intuition behind this design is that starting with a small kernel guides the training in a way that the first layers learn simpler more local features that fits in their window (kernel size).
Kernel sizes used in this experiment are 3 (a typical kernel size in the literature), and are increased to 5 (equal to the sampling frequency), 10, 15 and up to a size of 20.

Choosing a kernel size involves making a compromise between the propensity of a model to over-fitting or to over-generalising.  We limited the kernel size at 20 as we observed that any larger size would invariably lead to models that over-fit.

A similar judgement was made in the second section of the model (Recurrent layers). We ascertained that the `sweet spot' for hyper-parameters here was to use two GRU layers with 128 cells each. Their output was fed into a fully connected layer with 128 neurons with a leaky ReLU ($\alpha=0.3$) activation function \cite{maas2013rectifier} and finally to a dense layer with $N_s=25$ units with softmax activation.
We used Adam optimizer \cite{kingma2014adam} that could converge in 60-80 epochs, i.e. validation accuracy plateaued. The full architecture is shown in Figure~\ref{fig:model_arch}. Note that all the initial configurations discussed here are identified based on reported best practices in the literature and after some experimentation in our first case study. However, we also take a proper and standard approach for hyper-parameter tuning (Grid Search), that will be discussed in section ~\ref{sec:Methodology} (methodology) and reported in our research questions.

\subsection{Data Encoding}
Let's assume there are a total of Z flight scenarios in the test suite of a system. The input/output values of the black-box system create  one multivariate time-series ($T$) per flight scenario.  Each $T$ is defined as a tuple of $n$ univariate time series ($V_i$) of the same length of $T$ ($l_T$). Each $V_i$ corresponds to the recorded values for one of the inputs or outputs of the system:
\begin{equation} \label{eq:T_k}
    T = {\langle{V_1}, {V_2}, \ldots, {V_n}\rangle}
\end{equation}
\begin{equation} \label{eq:l_k}
    |{V_1}|=|{V_2}|=\ldots=|{V_n}|=l_T
\end{equation}

In practice, there will tend to be some lag between an input being presented to the system and the effect on the output signals. This impact may also play out over several time-steps. Since the system is a black-box, the precise extent of this lag is not known. Thus, we take both inputs and outputs as part of the time-series data to be fed as input into our deep learning models (as shown in Figure~\ref{fig:general_net}), where the states are predicted not just based on the current input/outputs, but rather using the current and past sequence of input/outputs. 

Including outputs from previous time steps is important because the externally observable outputs can be found important indicators of the current state of the UAV Autopilot system. As an example from our UAV case study, if the 'Elevator' outputs were not taken into account, a mid-flight ``descend'' state and the ``approach'' state right before landing would be indistinguishable if using the input sensor readings alone. 

The next task is to prepare the training data by labelling phases in the time-series with the corresponding state-label. Theoretically, this requires a domain expert to manually label each individual time step with a state ID. In practice, however, the experts do not need to manually label every single time step. They only need to identify the time step they believe the state has changed at, during a flight, which usually are not more than a handful, (on average 7 state changes happened during a typical test scenario in the Micropilot case and 22 changes in the Paparazzi's scenarios). 
This means that all time steps in between two consecutive state changes will be assigned labels, automatically. This reduces the manual work a lot and makes this process feasible. Note that we will also address the labeling cost in our transfer learning experiment. 

We encode the state information of the flight scenario ($T$) as a set of tuples ($CP_T$). In each tuple ($(t_{k}, s_i)$), $t_{k}$  refers to the time step where the system enters state $s_i$. The $CP_T$ may include repeated $s_i$s over different time steps. We denote the set of all possible states as $S$. 
\begin{equation}\label{eq:change_point}\begin{split}
    CP_T {}&{}= \big\{ (t_{k}, s_i), (t_{m}, s_j), \ldots,  \big\}\:, \textrm{where } \; s_i, s_j \in S \\
    N_s  {}&{}= |S|
\end{split}
\end{equation}

Note that, in this equation, $CP_T$ can include repeated states (e.g. where $i = j$) in different time steps (where $k \neq m$).
 
In summary, Each $T$ consists of $n$ univariate time series ($V_i$), one per observed I/O signal. Each $V_i$ includes exactly $l_T$ observations over time (each 200ms apart). The training data (the set of $Z$ flight scenarios ($T$)) is labeled with state IDs for every single time step. This data will be used as our state classification training set. These can be provided in the succinct tuple-format ($CP_T$), which reduces the labelling effort for the domain-expert. 


%

Different flight scenarios are not necessarily in the same length.
Therefore, to feed the data into TensorFlow, we first pre-process it to ensure that the signals are all of the same length ($L=18,000$), by zero-padding signals (adding zeros to the end of sequence) that are shorter than the longest signal. More details about this preprocessing can be found in the github repository of this paper\footref{foot:hybrid}.

\section{Empirical Evaluation} \label{sec:experiment}
The objective of this section is to evaluate our approach with the help of our industrial UAV system, provided by MicroPilot, as well as a well-known open source AutoPilot (Paparazzi). Our research questions are as follows:  
\subsection{Research Questions}

\begin{itemize}
    \item [RQ1] How effective is our proposed DNN-based technique on inferring states and state changes?
\end{itemize}

In order to answer this, we pose three sub-questions:

\begin{itemize}
    \item [RQ1.1] How does the proposed technique perform in detecting the state change points?\\ \emph{Does our approach outperform existing baseline approaches in terms of Precision, Recall and F1 scores?}
    \item [RQ1.2] How well does the proposed technique predict the internal state of the system?\\ \emph{How correct are the state-labels predicted by our approach?}
    \item [RQ1.3] How much does the proposed model owe its performance to being a hybrid model and how does it compare to a totally homogeneous model?\\ \emph{Would it suffice to just use either CNN or RNN?}
\end{itemize}

All of the above questions will be addressed in the context of our case study with our industrial partners at MicroPilot. However, this leads on to the second research question:

\begin{itemize}
    \item [RQ2] Do the results from the MicroPilot case study (RQ1) generalise to other UAV autopilot systems?
\end{itemize}

To answer this question we apply the technique to another UAV autopilot system; this time a non-proprietary open source one called: Paparazzi autopilot. To assess the generalisability of our approach, we pose the following sub-questions:

\begin{itemize}
    \item [RQ2.1] How does hyper-parameter tuning affect the generalizability of the results  on other UAV autopilot systems? \emph{Since DL performance is so sensitive to hyperparameter tuning, can the use of standard tuning approaches lead to a comparable performance on Paparazzi?}
    \item [RQ2.2] How well do the results on change-point detection generalise to other UAV autopilot systems? \emph{How good is our approach at detecting state-changes in Paparazzi?}
    \item [RQ2.3] How well do the results on state-labelling generalise to other UAV autopilot systems? \emph{How good is our approach at labelling the states in Paparazzi?}
\end{itemize}

Given that we use two similar systems (MicroPilot and Paparazzi) to answer RQ2, this raises the prospect of using Transfer Learning. We therefore also explore this as part of RQ2:

\begin{itemize}
    \item [RQ2.4] Is it possible to use Transfer Learning to use the model learned on one system to improve the performance on similar systems? \emph{Can we use the model from the MicroPilot study to improve the efficiency and accuracy of the learning process on Paparazzi?}
\end{itemize}

\subsection{Subject Systems}
In this section more details on how data was collected from each system will be presented.

\subsubsection{MicroPilot Autopilot}

MicroPilot's autopilot is a commercial autopilot with a codebase of 500k lines of C code. MicroPilot is the world-leader in professional UAV autopilot which develops both hardware and software for 1000+ clients (including NASA, Raytheon, and Northrop Grumman) in 85+ countries during the past 20+ years.

The primary control mechanism in the autopilot is a hierarchy of PID loops, as explained both in Section~\ref{sec:motivation} and \ref{sec:approach}.
High level commands are loaded on the autopilot as a flight plan. The flight plan looks like ``takeoff, climb to 300 feet, go to waypoint A, go to waypoint B, land''.
These commands determine which PID loops must be activated and what should their `desired values' be.
For example, a ``go to waypoint A'' command activates a PID loop that tries to minimize the distance between current location of the aircraft and point A. This is a high level loop that activates other lower-level PID loops to achieve its goal.
Those lower level loops can be one to maintain the altitude and another loop that keeps the aircraft heading on the straight line from current location to point A.
This hierarchical chain of `higher-level goals controlling lower level ones' goes on, down to the level that directly controls aerodynamic surfaces of the aircraft.

\begin{table}
    \caption{The collected I/Os of autopilot. The inputs are sensor readings and the outputs are the servo position update commands. All I/Os over time are used as the inputs of the state prediction model.}
    \label{tab:in_outs}
    \centering
\begin{tabularx}{\columnwidth}{lX}
                                                                                                                    \toprule
\multicolumn{2}{l}{\textbf{Inputs}}                                                                              \\ \midrule
Pitch     & The angle that aircraft's nose makes with the horizon around lateral axis                            \\
Roll      & The angle of aircraft's wings make with the horizon around longitudinal axis                         \\
Yaw       & The rotation angle of aircraft around the vertical axis                                              \\
Altitude  & AGL\footnotemark Altitude                                                                            \\
Air speed & Speed of the aircraft relative to the air                                                            \\ \midrule
\multicolumn{2}{l}{\textbf{Outputs}}                                                                             \\ \midrule
Elevator  & Control surfaces that control the Pitch                                                              \\
Aileron   & Control surfaces that control the Roll                                                               \\
Rudder    & Control surface that controls the Yaw                                                                \\
Throttle  & Controller of engine's power, ranges from 0 to 1                                                     \\
Flaps     & Surfaces of back of the wings that provide extra lift at low speeds, usually used during the landing \\ \bottomrule
\end{tabularx}
\end{table}
\footnotetext{Above Ground Level}

Control decisions in this software are made in a 5Hz loop, it means that every 200ms all the sensor inputs are read and based on the current state of the aircraft and the system's goal at the moment (e.g. maintaining a constant speed) decisions will be made and output is generated. Considering this, the best way to capture those data is in the end of each iteration of this loop. We inserted instrumentation code there, to log input and output values (listed in Table~\ref{tab:in_outs}) at the exact spot where they are updated. It is worth noting that the same data can be captured with a slightly elaborate wrapper around the interface to a purely black-box implementation.

\begin{figure}
    \centering
    \includegraphics[width=\columnwidth]{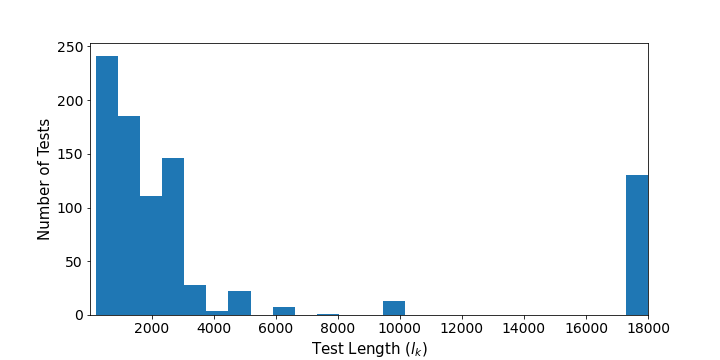}
    \caption{Distribution of flight log lengths for the $N=888$ (out of the original 948 available logs) logs that were kept in the dataset ($200 \leq l_T \leq 20,000$), from MicroPilot.}
    \label{fig:test_lengths}
\end{figure}

\paragraph{Test Scenarios}\label{sec:mp_test_scenarios}
MicroPilot has a repository of 948 system tests. We ran them in a software simulator\footnote{It is developed by MicroPilot and provides an accurate simulation of the aerodynamic forces on the aircraft, the physical environment irregularities (e.g. unexpected wind gusts), and noises in sensor readings} and collected the logged flight data, over time.
The test cases are system-level tests. Each test case includes a flight scenario for various supported aircraft. A flight scenario goes through up to 25 different phases in a flight such as ``take off'', ``climb'', ``cruise'', ``hitting way points'', and ``landing''.
Out of the 948 flight logs, we omitted 60 that were either too short or too long (shorter than 200 samples or longer than 20k samples). Figure~\ref{fig:test_lengths} shows the distribution of the remaining log lengths. After omitting those scenarios, the maximum length observed ($L$) was 18,000 samples.

The data items (test scenarios) were randomly split into three chunks of 90\%, 5\%, and 5\% for training, validation, and testing.

Note that separate test and validation sets are needed to facilitate proper hyper-parameters tuning, without generating data leakage.
The autopilot can be used in Software In The Loop (SWIL) and Hardware In The Loop (HWIL) modes \cite{melmoth2019true}. We used SWIL mode as it provides what was needed without any of the costs and hassles that come from HWIL mode.

\subsubsection{Paparazzi}\label{sec:paparazzi_data_collection}

\begin{figure}
    \centering
    \includegraphics[width=\columnwidth]{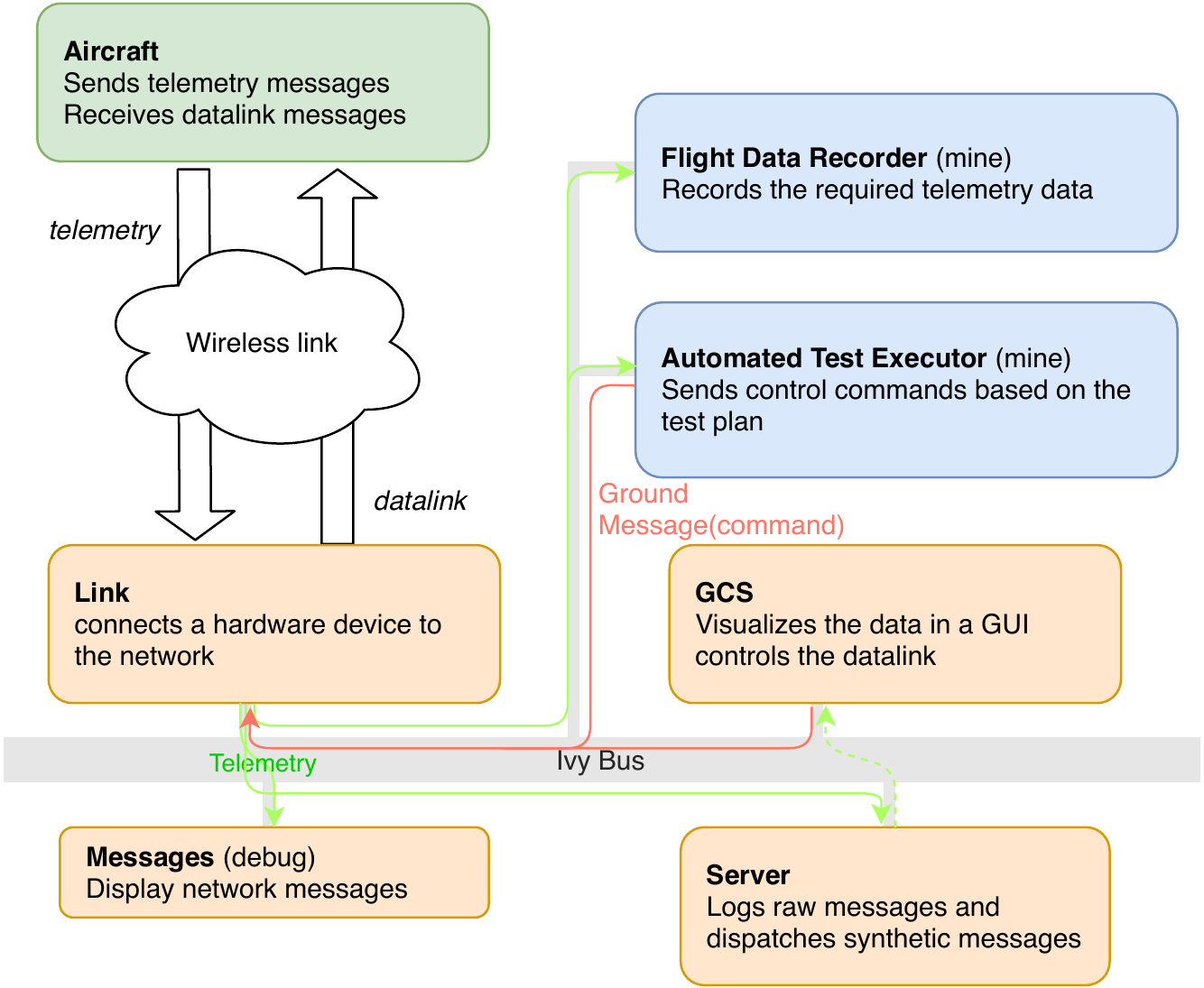}
    \caption{High level overview of communication links architecture in Paparazzi, including the components we added for this study's purpose. Diagram re-illustrated based on a diagram in \cite{hattenberger2014using} with some modifications.}
    \label{fig:paparazzi_comm_agents}
\end{figure}

Papparazi is a well-known and well-maintained autopilot software used in many studies with over 1100 stars on GitHub.
It provides a rich and flexible API that can be configured to record several different parameters in flight. The aircraft periodically sends data back to the ground station over a wireless link using a protocol called Paparazzi link. Paparazzi link is built over Ivy, a message bus protocol that uses UDP.
In Paparazzi's architecture a process called `link' interfaces the wireless link to the aircraft to the computer's network; on one side are the Paparazzi link messages that come and go as UDP datagrams and on its other side is the (often wireless\footnote{A wired connection is used in HWIL test mode as well as some scenarios where the autopilot equipment is used in a autonomous submarine, not an aircraft.}) connection to the aircraft.
In simulations, the modem and wireless communications are no longer needed, instead the autopilot runs as a separate process and mimics a wireless channel over the local network. (See Figure~\ref{fig:paparazzi_comm_agents})

One of the reasons for choosing Paparazzi over more recent open-source autopilots, in this study, is its efficient method for generating a dataset of flight logs, provided by the modular architecture.
Paparazzi comes with a multitude of small tools that could do most of what we needed in terms of instrumentation. There is a remote logger and a log player which are quite close to the instrumentation tool we need, however upon trying them in action, we figured that they cannot record some of the information that we need. Therefore, we developed a custom flight data recorder tool\footnote{\label{foot:pp}\urlA}.


\paragraph{Test Scenarios}
While MicroPilot had a large a number of system tests (in addition to other types of tests such as unit tests which we did not use), Paparazzi comes only with unit tests. This is possibly because it is not subject to the stringent certifications and approvals that commercial systems require.

To create tests for our study we created a fuzz testing tool that can automatically generate valid, diverse, and meaningful automated system tests for Paparazzi\footnote{During the development of our custom flight recorder and the testing tool we have been in contact with the Papparazi developers. They were very cooperative to include our implemented features for system test generation into their framework, which made our experiment possible. The helpful community was in fact another reason to choose Paparazzi as one of our case study subjects.}. 
Our tool takes an example flight-plan, automatically generates system tests and runs them in a simulator (or on hardware\footnote{Although we have not tested running tests on a hardware (HWIL) to confirm, but having implemented the protocol it potentially is capable of doing so.}), and also collects required telemetry data from the aircraft. The targeted randomizations in test inputs are augmented with the stochastic wind model in the simulation to further diversify the observed behaviours.
The tool (which we refer to as pprz\_tester) is available online \footref{foot:pp}. To make the logging and testing more similar to MicroPilot we added some patches to Paparazzi, for example increasing telemetry reporting rate from 2Hz to 5Hz. A list of these patches including the reason why that change was necessary or beneficial and the exact lines of code that need to be changed is available in the pprz-tester wiki\footref{foot:pp}.

The process of generating and running tests resulted in 378 runs (the final Paparazzi dataset size), where  a flight scenario comprises up to 20 different phases.
After collecting the data we performed some simple pre-processing steps to make them more similar to what the previous model was trained on. These pre-processing steps include normalizing some values as well as metric to imperial unit conversions. Full details are available in the Github repository of this paper\footref{foot:hybrid}. 

\begin{figure}
    \centering
    \includegraphics[width=\columnwidth]{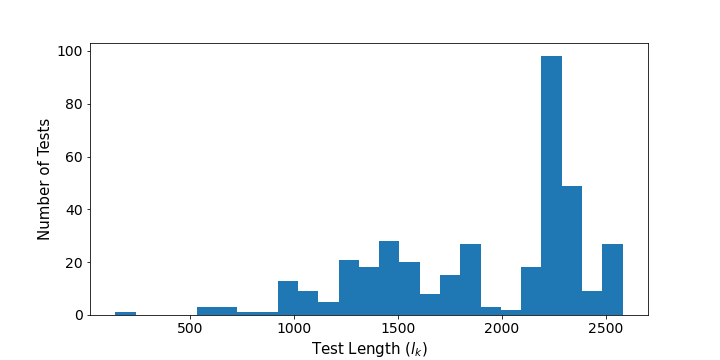}
    \caption{The histogram of number of tests per test lengths for Paparazzi dataset. The smallest tests contains 140 time steps worth of recorded flight data and the largest one has 2,580 time steps, equivalent of a 516-second flight (Sampling rate is 5Hz). }
    \label{fig:paparazzi_test_length}
\end{figure}

Figure~\ref{fig:paparazzi_test_length} shows the distribution of the test lengths. The test lengths range from 140 time steps (70 seconds) to 2,580 time steps, with a median of 2,170 and a mean of 1896. The dataset was split into 3 chunks after shuffling: 70\% of the data was used for training, 20\% was used as the validation set for tuning the hyper-parameters, and the remaining 10\% was set aside as the test data to measure the performance of the trained model.
Note that the splits are proportionally different to the chunks used for MicroPilot, since the dataset here is much smaller. In this case 5\% of the data, which is less than 20 flights, would not be sufficient for experimental analysis.

\subsection{Methodology}
\label{sec:Methodology}
\subsubsection{RQ1: Measuring CPD Accuracy}
\label{sec:CPD_metrics}

We start from a labelled data-set, where for a given multivariate time-series there is a corresponding series of state-labels. For this we use the simulator to collect the exact time a state-change happens and the actual state labels (ground truth).

Given that in this task there is an inherent class imbalance (there are far more points where a change has \textit{not} happened compared to points with a state-change positive label), we avoid using the proportional measure of accuracy and report both Precision and Recall.
The original Precision / Recall metrics require some modification, in order to accommodate a degree of approximation around the exact time stamp at which a state-change has happened. To handle this, similar to related work \cite{Truong2018ChangePointSurvey, Lee2018TimeSeriesSegmentation}, we use a tolerance margin $\tau$. If a detected state-change ($\in\!\hat{C\!P}_T$) is within $\pm\tau$ of a true change ($\in\!\!{C\!P}_T$), we call the prediction a True Positive, otherwise it is a False Positive. Similar adjustment to definition is applied for True Negative and False Negative.
Assume $\hat{O}$ is the output of model $\mathcal{F}$, as follows (the $\hat{}$ notations in these equations refer to the predicted values):
\begin{equation}\label{eq:model_F}\begin{split}
    \mathcal{F}(T) {}&{} \colon \mathbb{R}^{L\times n}\to S^{L} \\
    \hat{O} ={}&{} [\hat{s}_i \in S] ^L_{i=1} = \mathcal{F}([ V_1 \: V_2 \; \ldots \; V_n]) \\
\end{split}
\end{equation}
We denote change points for trace element $T$ as:
\begin{equation} \label{eq:cp_hat}\begin{split}
    \hat{C\!P}_T = \big\{(\hat{t}, \hat{s}_{\hat{t}})\: |\: \hat{s}_{\hat{t}} \neq \hat{s}_{\hat{t}-1} \big\}
\end{split}
\end{equation}
Here, $\hat{s}_t$ refers to $\hat{t}$-th element of output vector $\hat{O}$. Based on this the confusion matrix elements are calculated as follows:

\begin{equation} \label{eq:metrics}
\begin{split}
T\!P ={}&{}\Big|\big\{ (\hat{t}, \hat{s}_{\hat{t}}) \in \hat{C\!P}_T \;\big|\; \exists\: (t, s_t) \in C\!P_T \;\text{s.t.}\; |t - \hat{t}| < \tau\big\}\Big| \\
F\!P ={}&{}\Big|\big\{ (\hat{t}, \hat{s}_{\hat{t}}) \in \hat{C\!P}_T \;\big|\; \nexists\: (t, s_t) \in C\!P_T \;\text{s.t.}\; |t - \hat{t}| < \tau\big\}\Big| \\
F\!N ={}&{}\Big|\big\{ (t, s_t) \in C\!P_T \;\big|\; \nexists\: (\hat{t}, \hat{s}_{\hat{t}}) \in \hat{C\!P}_T \;\text{s.t.}\; |t - \hat{t}| < \tau\big\}\Big|
\end{split}
\end{equation}

From these we calculate Precision, Recall and F1 in the usual way: $Precision = \frac{TP}{TP+FP}$. $Recall = \frac{TP}{TP+FN}$, and $F1 = \frac{2 * TP}{2 * TP + FP + FN}$.

We calculate these measurements with three values for $\tau$: 1, 3, 5 seconds equivalent to 5, 15, 25, at 5Hz sampling rate. The smaller the tolerance, the stricter the definitions become (leading to reduced accuracy scores).

For baselines for RQ1.1 we used `ruptures' library developed by authors of a recent CPD survey study \cite{Truong2018ChangePointSurvey}. It is a library to facilitate experimentation with CPD algorithms, and provides a variety of search and cost functions, which can be combined to form different state-of-the-art algorithms.

There are three search methods implemented which were suitable to use as baselines in our experiments, i.e. that do not require assumptions about the number of change points, distribution, etc. Pelt \cite{killick2012optimal} is the most efficient exact search method. Two other ones are approximate search methods: bottom-up segmentation and window-based search method.
After trying to run Pelt  on MicroPilot's data, we realized that it is prohibitively time-consuming compared to the approximate methods without providing much better results. As a result we restrict ourselves to the bottom-up and the window-based segmentation methods for baselines.

For the cost function, we tried ``Least Absolute Deviation'', ``Least Squared Deviation'', ``Gaussian Process Change'', ``Kernelized Mean Change'', ``Linear Model Change'', ``Rank-based Cost Function'', and ``Auto-regressive model change'' as defined in the library. Their parameters were left as default.
To optimize the number of change points a penalty value (linearly proportionate to the number of detected change points) is added to the cost function, which limits the number of detected change points, the higher the penalty the fewer reported change points. We tried three different values (100, 500, and 1000) for the penalty parameter.

Classical change point detection algorithms often leverage statistical measures to detect a shift in the sequence. These methods assume local independence in their multivariate sequence, which makes the problem easier to solve.  We used the Spearman's rank correlation coefficient to measure local dependency in our datasets.  Most variables have no/weak pair-wise correlation ranging between 0 to 0.28. We only found a moderate correlation between Air speed and Elevator and between Aileron and Rudder with coefficients ranging between 0.56 and 0.74. With further observation, we found these moderate correlations are due to these variables being 0 for long periods of time. Thus, it is safe to assume that local dependence in our datasets is weak.

In RQ1.2 (checking that the label of the predicted state matches the actual state) we have a multi-class classification problem. For this we calculate one set of Precision / Recall values per class (state label). We then report the mean value across all classes.

As a baseline for RQ1.2 we used Scikit-learn's implementation of the classification algorithms: A ridge classifier (Logistic regression with L2 regularization) and three decision trees. The ridge classifier was configured to use the built-in cross validation to automatically chose the best regularization hyper-parameter $\alpha$ in the range of $10^{-6}$ to $10^6$. Each decision tree was regularized by setting ``maximum number of features'' and ``maximum depth''.

To prepare the data to be fed to the classification algorithms, we used a sliding window of width $w$ over the 10 time-series values and then flattened it to make a vector of size $10w$ as the features. For the labels,
which are categorical values for different states, we used their one-hot encoded representation (i.e., rather than having one integer feature with state IDs between 1 to 25, we use 25 binary features, with each encoding corresponding to a particular state). 

The window sizes were chosen as same as the sizes of convolutional layers' kernel sizes (3, 5, 10, 15, 20), to make the baselines better comparable with our method.

The ridge classifier was configured to use the built-in cross validation to automatically chose the best regularization hyper-parameter $\alpha$.
To regularize the decision trees we tried: no limits, $\sqrt{10w}$, and $\log_2{10w}$ for ``maximum number of features'' regularization parameter. To find best ``maximum depth'' we first tried having no upper bound and observed how deep the tree grows; then we tried multiple numbers less than the maximum, until a drop in performance was observed.

To answer RQ1.3, we set up four new variations of the model. The first two variations (the CNN-only and RNN-only models) are meant for an ablation study, where in the CNN-only model the GRU layers are dropped and in the RNN-only model the Conv layers are removed (the remaining of the model is unchanged). The idea is to assess which of these two components is more important and contributes more in the hybrid model. We also look at two other variants (RNN-Full and CNN-Full), as baselines. The RNN-Full architecture includes 2 GRU layers with cell size of 200 and CNN-Full on the other hand includes more convolutional layers each with more number of filters compared to the hybrid model. Both RNN-Full and CNN-Full models have the same size parameter-wise as the hybrid model to make them a fair comparison.(Around 400k parameters each)


\subsubsection{RQ2: Establishing the generalisation to other UAV autopilot systems}

For RQ2 we use the same basic performance measurements of Precision, Recall and F1 (and their multi-class variants) as covered in Section \ref{sec:CPD_metrics}.

For RQ2.1, we designed a model creation and evaluation pipeline that takes hyper-parameters as the input and outputs the model performance scores on test data (tuning data) as its output. The hyper-parameters that we searched over are:
\begin{itemize}
    \item Number of GRU layers: 1 or 2
    \item Number of GRU cells in the recurrent section: 5 values between 64 and 512
    \item Number of convolutional filters in each layer: 5 values between 16 to 72
    \item The size of convolution kernels and the number of convolutional layers: 3-6 layers with increasing kernel size, starting from kernel sizes 3 or 5
    \item The learning rate of Adam optimizer: 3 values from $3\times 10^{-4}$ to $3\times10^{-3}$
\end{itemize}

The model architecture is shown in Figure~\ref{fig:model_arch}.
We performed a grid search over these parameters, using Tensor Board\footnote{\url{https://www.tensorflow.org/tensorboard}} to track the metrics and determine the right balance.
Tensor Board is a monitoring tool made for TensorFlow that provides great insight for better training TensorFlow models.
In total, there were 520 configurations that were used to train models on the training data and evaluated on test (tuning) data.

For RQ2.2 we re-applied the same methodology as described in RQ1.1 (evaluating the CPD accuracy on MicroPilot), but applied it to the Paparazzi system. Since the Paparazzi dataset was smaller, it became feasible (though still really time-consuming) to try ``Pelt'' as well.
When using the window-based search method, we left the window size parameter at the default size of 100 \cite{keogh2001online}.

In RQ2.3 we again used the same configurations and procedures as the MicroPilot's case (RQ1.2). The only difference was the removal of the depth limit from the decision trees (since this activity had turned out to have a negligible effect on accuracy, as shown in the results for RQ1.2 later on).

For RQ2.4, the hybrid model developed for MicroPilot is used as a source model. We compare the predictions that the source model (tailored for and trained on the Paparazzi dataset) provides against the predictions made by a transfer learning technique based on the source model.

To see the effect of transfer learning, we apply it on the Paparazzi case study. Recall that Paparazzi had  a very small set of tests and we had to generate tests using our own Fuzz testing tool. This, therefore, presents a good example of a scenario that suits transfer learning. The objective of this sub-RQ is to study the extent to which transfer learning is effective in this setting. 

To apply the transfer learning to our source model from MicroPilot, we use the last two fully connected layers for fine tuning and freeze the remaining layers. We allocate a very small subset of the Paparazzi's whole dataset for training (only 5 test cases -- less than 2\% of the whole dataset). The rest of the data is used as test sets. We use the same data points to train and test the source model, where all layers are trained from scratch (initialized with random weights) on the small training set, without freezing any layers; i.e., no transfer learning. We then calculate the mean for each metric over all the generated results in a K-fold cross-validation (with only 5 items as training set in each fold). 

\subsubsection{Experiment Execution Environment} \label{sec:machines_config}
Training and evaluation of the deep learning model was done on a single node running Ubuntu 18.04 LTS (Linux 5.3.0) equipped with Intel Core i7-9700 CPU, 32 gigabytes of main memory, and 8 gigabytes of GPU memory on a NVIDIA GeForce RTX 2080 graphics card.
The code was implemented using Keras on TensorFlow 2.0 \footnote{https://www.tensorflow.org/}.

The baseline models could not fit on that machine, so two nodes on Compute Canada's Beluga cluster, one with 6 CPUs and 75GiB of memory and one with 16 CPUs and 64GiB of memory, were used to train and evaluate them.

In this environment we have collected the execution costs of each technique in terms of actual time, and informally discuss them in the results for each RQ as well.

\subsection{Results} \label{sec:results}
In this section, we present the results of the experiments and answer the two research questions.
\subsubsection{RQ1.1 - CPD Performance}

\begin{table*}
\caption{Change point detection precision, recall, and F1-score calculated for the baseline methods using three values of tolerance ($\tau$) for multiple configurations.}
\label{tab:rq1-results}
\resizebox{\textwidth}{!}{%
\begin{tabular}{llcccccccccc}
\toprule
  \textbf{Cost Function} &
  \textbf{Search Method} &
  \textbf{Penalty} &
  \textbf{Prec.} &
  \textbf{Recall} &
  \textbf{F1} &
  \textbf{Prec.} &
  \textbf{Recall} &
  \textbf{F1} &
  \textbf{Prec.} &
  \textbf{Recall} &
  \textbf{F1} \\
                                                  &              &   & \multicolumn{3}{c}{$\tau=1$s} & \multicolumn{3}{c}{$\tau=3$s} & \multicolumn{3}{c}{$\tau=5$s} \\  \toprule
\multirow{2}{*}{\textbf{Autoregressive Model}}     & Bottom Up    & 1000 & 10.43\%    & 75.44\%    & 18.33\%   & 21.21\%    & 80.32\%    & 33.55\%   & 28.94\%    & 81.22\%    & 42.68\%   \\
                                                   & Window Based & 100  & 2.94\%     & 3.98\%     & 3.38\%    & 8.53\%     & 11.41\%    & 9.76\%    & 12.89\%    & 17.54\%    & 14.86\%   \\ \midrule
\multirow{2}{*}{\textbf{Least Absolute Deviation}} & Bottom Up    & 500  & 7.32\%     & 52.54\%    & 12.85\%   & 17.52\%    & 87.73\%    & 29.20\%   & 25.02\%    & 88.95\%    & 39.05\%   \\
                                                   & Window Based & 500  & 5.24\%     & 8.31\%     & 6.42\%    & 15.20\%    & 24.03\%    & 18.62\%   & 21.79\%    & 38.25\%    & 27.76\%   \\ \midrule
\multirow{2}{*}{\textbf{Least Squared Deviation}}  & Bottom Up    & 1000 & 7.44\%     & 85.09\%    & 13.68\%   & 16.40\%    & 89.81\%    & 27.74\%   & 24.16\%    & 90.47\%    & 38.14\%   \\
                                                   & Window Based & 500  & 3.59\%     & 6.79\%     & 4.70\%    & 10.27\%    & 16.51\%    & 12.66\%   & 16.18\%    & 26.84\%    & 20.19\%   \\ \midrule
\multirow{2}{*}{\textbf{Linear Model Change}}      & Bottom Up    & 100  & \textbf{37.59\%}    & 28.98\%    & \textbf{32.73\%}   & \textbf{45.20\%}    & 38.39\%    & \textbf{41.52\%}   & \textbf{48.07\%}    & 41.36\%    & \textbf{44.46\% }  \\
                                                   & Window Based & 500  & 6.70\%     & 4.14\%     & 5.12\%    & 20.50\%    & 13.05\%    & 15.95\%   & 38.78\%    & 26.77\%    & 31.67\%   \\ \midrule
\multirow{2}{*}{\textbf{Gaussian Process Change}}  & Bottom Up    & 100  & 3.77\%     & \textbf{92.23\%}    & 7.25\%    & 8.99\%     & \textbf{92.23\%}    & 16.39\%   & 13.53\%    & \textbf{92.23\%}    & 23.60\%   \\
                                                   & Window Based & 100  & 2.94\%     & 3.95\%     & 3.37\%    & 8.69\%     & 11.50\%    & 9.90\%    & 13.64\%    & 18.30\%    & 15.63\%   \\ \midrule
\multirow{2}{*}{\textbf{Rank-based Cost Function}} & Bottom Up    & 100  & 13.45\%    & 60.19\%    & 21.98\%   & 19.49\%    & 80.10\%    & 31.35\%   & 22.98\%    & 87.23\%    & 36.38\%   \\
                                                   & Window Based & 100  & 8.10\%     & 13.70\%    & 10.18\%   & 15.72\%    & 30.73\%    & 20.80\%   & 21.38\%    & 46.64\%    & 29.32\%   \\ \midrule
\multirow{2}{*}{\textbf{Kernelized Mean Change}}   & Bottom Up    & 100  & 4.13\%     & 3.24\%     & 3.63\%    & 12.22\%    & 8.14\%     & 9.77\%    & 15.38\%    & 10.58\%    & 12.54\%   \\
                                                   & Window Based & 100  & 2.82\%     & 3.00\%     & 2.91\%    & 10.14\%    & 8.40\%     & 9.19\%    & 13.64\%    & 12.61\%    & 13.10\%   \\ \midrule \midrule
\textbf{Our Approach} & \multicolumn{2}{l}{\textbf{-}}
& \textbf{56.77\%} &	79.32\% &	\textbf{66.18\%} & \textbf{69.58\%} &	88.88\% &	\textbf{78.06\%} & \textbf{79.82\%} &	91.87\%	&   \textbf{85.42\%} \\\bottomrule

\end{tabular}%
}
\end{table*}

Table~\ref{tab:rq1-results} shows the results of running CPD algorithms for various configurations 
. For each search method and cost function pair only one of the penalty values which resulted in the highest F1 scores for all $\tau$ values is reported.

As is to be expected, larger values of $\tau$ lead to improved scores.
Another observation is that the bottom-up segmentation consistently outperforms the window-based segmentation method. We can also see that the linear cost function beats all the other ones in terms of precision. The Gaussian cost function achieves much higher recall values, but at the expense of precision. This cost function results in the detection of numerous change points spread across the time axis, so there is a good chance of having at least one change point predicted close to each true change point (hence the high recall), but also there are a lot of false positives, which leads to a low precision.
Our approach (see the last line in the same table) shows improved scores throughout. Its results almost double the F1 score of the best performing baseline.

In terms of execution cost, running all 42 different settings of CPD algorithms on the whole dataset took a bit over 12 hours in the cloud using 16 CPUs and 64GB of main memory. The deep learning model on the other hand takes about an hour to train (which only needs to be done once), on a smaller machine (see section~\ref{sec:machines_config}). It made predictions on the whole dataset in less than a minute.

So to answer RQ1, our method has shown $(66.18/32.73)-1=102.20\%$ improvement in F1 score with $\tau=1s$, $(78.06/41.52)-1=88.00\%$ with $\tau=3s$, and $(85.42/44.46)-1=92.13\%$ with $\tau=5s$; almost doubling the score compared to the baselines.

\begin{rqanswer}
The proposed model, which requires less memory compared to traditional CPD algorithms, improved their best performance by up to 102\%, measured by F1 score, in less execution time.
\end{rqanswer}

\subsubsection{RQ1.2 - Multi-class Classification Performance}
To answer RQ1.2, we first compare different configurations of the baseline methods using the F1 score (harmonic mean of precision and recall) on the test data. The results are presented in Table~\ref{tab:rq2-1-results}.

\begin{table}
\caption{Precision, recall, and F1 score of ridge classifiers (linear classifiers with L2 regularization) and decision tree classifiers (DT) with different sliding window widths ($w$). For each algorithm on each $w$ several hyper-parameters were applied producing 152 different models. In this table, we only show the results of the best performing model in each group.}
\label{tab:rq2-1-results}
\resizebox{\linewidth}{!}{%
\begin{tabular}{clcclll}
\toprule
\textbf{w} &
  \multicolumn{1}{c}{\textbf{Classifier}} &
  \multicolumn{1}{c}{\textbf{\begin{tabular}[c]{@{}c@{}}Max\\ Depth\end{tabular}}} &
  \multicolumn{1}{c}{\textbf{\begin{tabular}[c]{@{}c@{}}Max\\ Features\end{tabular}}} &
  \multicolumn{1}{c}{\textbf{Prec.}} &
  \multicolumn{1}{c}{\textbf{Recall}} &
  \multicolumn{1}{c}{\textbf{F1}} \\ \midrule
3  & Ridge & -   & -          & 71.39\% & 20.73\% & 32.13\% \\
3  & DT    & -   & -          & 69.21\% & 82.36\% & 75.21\% \\ \midrule
5  & Ridge & -   & -          & 69.15\% & 21.89\% & 33.26\% \\
5  & DT    & 100 & -          & 68.37\% & \textbf{83.16\%} & 75.04\% \\ \midrule
10 & Ridge & -   & -          & 71.97\% & 24.02\% & 36.02\% \\
10 & DT    & 260 & -          & 67.94\% & 79.14\% & 73.12\% \\ \midrule
15 & Ridge & -   & -          & 76.87 & 25.90\% & 38.75\% \\
15 & DT    & -   & $\sqrt{10w}$ & 69.06\% & 80.76\% & 74.45\% \\ \midrule
20 & Ridge & -   & -          & \textbf{80.38\%} & 26.50\% & 39.86\% \\
20 & DT    & 175 & $\sqrt{10w}$ & 73.21\% & 82.16\% & \textbf{77.42\%} \\  \midrule
   & \multicolumn{3}{l}{\textbf{Our Approach}} & \textbf{86.29\%} & \textbf{95.04\%} & \textbf{90.45\%} \\
\bottomrule
\end{tabular}%
}
\end{table}

Comparing the baseline methods with our approach (the last row) in Table~\ref{tab:rq2-1-results} shows that our model outperforms all baselines. Comparing it with the model with the best F1-score shows a $(86.29/73.21)-1=17.87\%$ improvement in precision as well as a $(95.04/82.16)-1=15.68\%$ improvement in recall that means $(90.45/77.42)-1=16.83\%$ overall improvement in F1-score.

\begin{figure*}
    \centering
    \includegraphics[width=\textwidth]{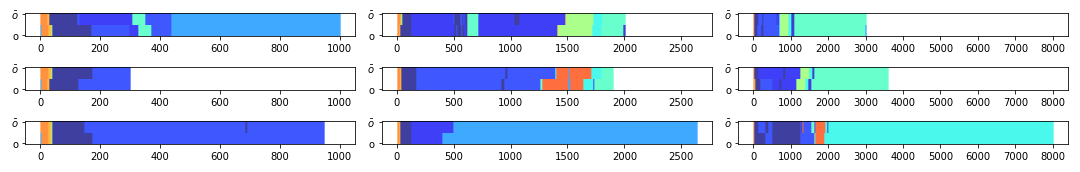}
    \caption{Evaluation of the model on 9 random test data. Each graph shows the states in one run of the system. The colors show the states. The top-half of each plot depicts model's prediction of the system states ($\hat{O}$) and the bottom-half shows the true labels($O$). Each column contains 3 samples from different flight lengths consisting of 3 groups, short($|l|$$<$1760 time-steps; the first tertile), medium(1760\textless$|l|$\textless2810; the second tertile) and long(2810\textless $|l|$). The x-axis shows the time steps, each equal to 200ms.}
    \label{fig:test_0}
\end{figure*}

To get a visual impression of how good our predictions are in practice, similar to what was explained in Figure~\ref{fig:states}, Figure~\ref{fig:test_0} shows the output of our model alongside the ground truth. The horizontal axis shows sample ID (time) and the states are color coded; each bar is split horizontally between ground-truth and predicted state values. This indicates that the algorithm performs better when the state changes are farther apart. There are some state changes that happen in quick succession which are not detected. 
This might happen because very quick frequent changes may be considered as noise, and not a pattern to infer. Whereas the baseline models only see one window of the data at a time, convolutional layers are more generalized and flexible since each filter in each layer is comparable to a sliding window. 

As shown in Table~\ref{tab:rq2-1-results}, a larger window size leads to an improved performance. However, the downside of this is that it becomes significantly more difficult to train a model with large window sizes. In addition, convolutions can automatically learn preprocessing steps that could be beneficial such as a moving average. Each convolutional filter can learn a linear combination of its inputs. So when the convolutional layers are stacked on each other with non-linear activation functions in between, the hypothesis space they can learn grows, probably much larger than most of the baseline algorithms here. 
The fact that the performance improves as the window size increases is probably due to the ability of the  recurrent cells (such as GRU) to capture long-term dependencies that do not necessarily fall into one window. 

In terms of the training complexity (time and memory), our approach is much less resource-consuming. This can largely be attributed to the use of deep learning. In baseline models, as the window size $w$ grows the training and evaluation complexity also grows, up to a point where they ran out of memory. This forced us to train them in the cloud. Meanwhile, as mentioned earlier, the deep learning model could be trained on a 8GB GPU in roughly an hour.

\begin{rqanswer}
The proposed model requires less than half as many CPUs and 70\% as much memory compared to the best performing classical ML model. However, it improved upon the best baseline performance by up to 17\% (in terms of F1 score) and required significantly less execution time.
\end{rqanswer}


\subsubsection{RQ1.3 - Ablation study and Hybrid vs. Homogeneous models}
As the results in Table~\ref{tab:rq5} suggest\footnote{Note that the results in the last column differ (around 1\% in absolute value) from their corresponding results in tables~\ref{tab:rq2-1-results} and \ref{tab:rq1-results} due to randomizations in splitting the data into training, testing, and validation sets.}, compared to the CNN-only model, the RNN-only results were closer to those of the full hybrid model, reaffirming the important role GRU units play in capturing long-term relations in the data and inferring the system's internal state.

The hybrid architecture consistently outperforms the fully convolutional (CNN-Full) and fully recurrent (RNN-Full) baselines. This confirms our hypothesis that although the RNN units are especially important, the CNN layers offer a complementary set of functionalities, resulting in a superior hybrid model performance.

\begin{table}
    \centering
    \caption{Comparing the hybrid model's performance (precision, recall, F1) with its sub-models (CNN-only and RNN-only), as well as, two comparative homogeneous models (RNN-Full and CNN-Full).}
    \label{tab:rq5}
    \begin{tabular}{llcccc}
        \toprule{}
         &  \textbf{RNN}  &  \textbf{CNN}  &  \textbf{RNN}  &  \textbf{CNN} &  \textbf{Hybrid} \\
         {}
         &  \textbf{only}  &  \textbf{only}  &  \textbf{Full}  &  \textbf{Full} &  \textbf{Model}\\
        \midrule
        Prec.(1s) &         45.31\% &    38.12\% & 47.12\% &          46.72\% & \textbf{53.84\%} \\
        Recall(1s)    &         60.56\% &    58.50\% & 55.62\% &          50.50\% & \textbf{67.94\%} \\
        F1(1s)        &         51.84\% &    46.16\% & 50.97\% &          48.53\% & \textbf{60.06\%} \\ \midrule
        Prec.(3s) &         56.06\% &    50.97\% & 57.31\% &          63.47\% & \textbf{72.00\%} \\
        Recall(3s)    &         78.00\% &    69.12\% & 68.44\% &          65.19\% & \textbf{88.56\%} \\
        F1(3s)        &         65.25\% &    58.69\% & 62.41\% &          64.31\% & \textbf{79.44\%} \\ \midrule
        Prec.(5s) &         68.75\% &    67.38\% & 74.94\% &          70.56\% &   \textbf{78.56\%} \\
        Recall(5s)    &         81.81\% &    73.75\% &  85.50\% &          78.81\% & \textbf{93.75\%} \\
        F1(5s)        &         74.69\% &    70.44\% & 77.12\% &          85.06\% & \textbf{85.50\%} \\ \midrule
\multicolumn{1}{l}{Class Prec.} &    81.56\% & 71.56\% &71.88\% &          80.81\% & \textbf{88.44\%} \\
\multicolumn{1}{l}{Class Recall}&    91.88\% & 86.88\% & 83.12\% &          89.75\% & \textbf{94.50\%} \\
\multicolumn{1}{l}{Class F1}    &    86.44\% & 78.44\% & 77.12\% &          85.06\% & \textbf{91.38\%} \\
        \bottomrule
        \end{tabular}
\end{table}

\begin{rqanswer}
The hybrid architecture significantly outperforms a comparable RNN model (RNN-Full) or fully convolutional model (CNN-Full). The recurrent section (RNN-only) plays a more important role in the model's performance compared to the convolutional units (CNN-only).
\end{rqanswer}

\subsubsection{RQ2.1 - Hyper-parameter Tuning}

We used Tensor Board to compare the effects of different hyper-parameters on the model's performance (across 520 configurations in our grid search). We looked at 6 metrics to tune (precision and recall for CPD with $\tau=1s$ and $\tau=5s$ and the precision and recall for state classification). 

The tuning results showed that a) the number of GRU cells has a high correlation with both CPD precision and classification precision, b) the learning rate has the largest correlation (absolute) with the performance metrics, and c) the next most important contributing factor to the reported metrics was the convolutional layer counts and sizes.


Based on the previous studies, especially in the field of computer vision, we can say that each filter learns one feature so it is more effective to have small filters that are easier to train and can learn multiple features of the data.

\begin{figure*}
    \centering
    \includegraphics[width=\textwidth]{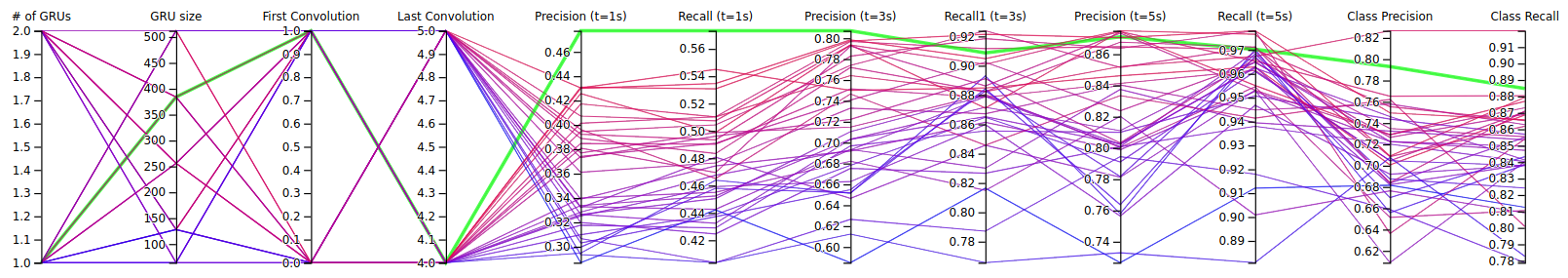}
    \caption{Parallel coordinates view showing the effects of selected hyper-parameters on the performance measures.}
    \label{fig:precision_recall_parallel_coordinates}
\end{figure*}

Figure ~\ref{fig:precision_recall_parallel_coordinates} summarizes the tuning results, after filtering out some of the poor configurations (those with $<40\%$ CPD recall (t=1s)).
The figure shows 3 + 8 parallel axes (3 hyper-parameters and 8 metrics).
The green highlighted line is the best configuration with the highest area under the curve. 
The best configuration uses 384 GRU cells in 1 layer, 16 filters per convolutional layer with sizes ranging from 5 to 15, and a slow learning rate of $3\times10^{-4}$.

To conclude the tuning study, Table~\ref{tab:results_no_tuning} reports the results of the model evaluation with and without (i.e., using the same model parameters as the MicroPilot case study's model parameters) tuning. The tuning has increased all evaluation scores (in the range of 11.5\% to 42.5\%). For instance, looking at the four F1 scores, we see an improvement between 18.3\% (from 79.11\% to 93.59\%, for the CPD's F1 with $\tau=5s$) to 34.8\% (from 64.87\% to 87.45\%, for the CPD's F1 with $\tau=3s$). This has brought the Paparazzi results into line with the MicroPilot case study. 

\begin{table}[]
\caption{The results of our model trained on the Paparazzi dataset before hyper parameter tuning (i.e. using the hyper-parameters that worked best for MP on Paparazzi as well) vs. after fine-tuning hyper-parameters on the Paparazzi's dataset.}
\label{tab:results_no_tuning}
    \centering
\resizebox{\linewidth}{!}{%
    \begin{tabular}{lcc}
    \toprule
\textbf{Evaluation Metric} & \textbf{Default Hyper-params} & \textbf{Tuned Hyper-params} \\\midrule
Precision ($\tau=1s$)      &  28.90\%  &  49.29\% \\
Recall ($\tau=1s$)         &  37.44\%  &  56.69\% \\
F1 ($\tau=1s$)             &  32.62\%  &  52.78\% \\ \midrule
Precision ($\tau=3s$)      &  59.34\%  &  85.50\% \\
Recall ($\tau=3s$)         &  71.52\%  &  90.28\% \\
F1 ($\tau=3s$)             &  64.87\%  &  87.84\% \\ \midrule
Precision ($\tau=5s$)      &  73.85\%  &  91.65\% \\
Recall ($\tau=5s$)         &  85.18\%  &  95.90\% \\
F1 ($\tau=5s$)             &  79.11\%  &  93.75\% \\ \midrule
State Detection Prec.  &  57.35\%  &  80.86\% \\
State Detection Recall     &  78.90\%  &  88.23\% \\
State Detection F1         &  66.41\%  &  84.38\% \\
\bottomrule
    \end{tabular}%
}
\end{table}
\begin{rqanswer}
Hyper-parameter tuning is crucial for the generalisability of our approach. The tuned model improves the evaluation scores between 9\% to 25\%.
\end{rqanswer}

\subsubsection{RQ2.2 - CPD Performance on Paparazzi}
The results of applying baseline algorithms on the Paparazzi dataset are visualised in Table~\ref{tab:cpd_paparazzi}.
The range of CPD performance varies across the techniques. 
The setting that achieved the best result is the Pelt algorithm using an L1 cost function and a high penalty coefficient of 1,000. However, Pelt is a slow algorithm that would rapidly become infeasible to run on a larger dataset, as it was the case for the MicroPilot case study.
In this Paparazzi case, running Pelt took more than 14 hours on the same machine that performed all other CPD algorithms in less than 1 hour. What you see in table~\ref{tab:cpd_paparazzi} is the summary of the results of more than 23,800 experiments.

The baseline results can be contrasted with the neural network results in the last three rows of the table. This shows improvements of 47.88\%, 34.81\%, and 18.30\% in F1 scores compared to the best results in the baselines (for $\tau = 1, 3, 5$ seconds respectively). Although this model still performed better than the baselines, the improvement margin was lower than in RQ1.1. 
This is not due to any decrease in performance by our approach, but by a relative increase in performance by the baselines on Paparazzi's data.
We posit that this is because the Paparazzi dataset is very small compared to MP; Paparazzi  consists of ~300 tests, each containing up to 2,500 samples vs. ~900 tests in MP with series comprising up to 18,000 samples.

\begin{rqanswer}
The proposed approach showed a near 48\% improvement over the baselines with a 94\% F1 score, confirming that it is a feasible approach even with smaller data.

\end{rqanswer}

\begin{table*}[t]
\caption{Change point detection methods performance on Paparazzi dataset. Since the 100\% recalls are actually outliers, the next largest recall values are in bold face as well.}
\label{tab:cpd_paparazzi}
    \centering
\resizebox{\textwidth}{!}{%
\begin{tabular}{llcccc|ccc|ccc}
\toprule
  \textbf{Cost Function} &
  \textbf{Search Method} &
  \textbf{Penalty} &
  \textbf{Prec.} &
  \textbf{Recall} &
  \textbf{F1} &
  \textbf{Prec.} &
  \textbf{Recall} &
  \textbf{F1} &
  \textbf{Prec.} &
  \textbf{Recall} &
  \textbf{F1} \\
                                                  &              &   & \multicolumn{3}{c}{$\tau=1$s} & \multicolumn{3}{c}{$\tau=3$s} & \multicolumn{3}{c}{$\tau=5$s} \\  \toprule
\multirow{3}{*}{\textbf{Autoregressive Model}}
    & Bottom Up    & 500  & 13.65\% &    14.13\% & 18.89\% & 39.73\% &     36.40\% & 36.14\% & 57.11\% &     53.20\% & 50.04\% \\
    & Pelt         & 500  & 13.56\% &    13.85\% & 19.03\% & 39.67\% &     36.17\% & 36.03\% & 56.97\% &     52.89\% & 49.83\% \\
    & Window Based & 100  & 15.97\% &     7.59\% & 13.54\% & 45.47\% &     20.40\% & 29.56\% & 64.16\% &     30.76\% & 41.20\% \\ \midrule
\multirow{3}{*}{\textbf{Least Absolute Deviation}}
    & Window Based & 100  & 24.50\% &    15.92\% & 19.85\% & 56.39\% &     37.24\% & 44.53\% &        68.94\% &     48.20\% & 56.23\% \\
    & Pelt         & 1000 &\textbf{26.74\%}&34.08\% & 29.92\% &\textbf{60.77\%}&     77.32\% &\textbf{67.74\%}&\textbf{74.73\%}&     93.16\% &\textbf{82.61\%}\\
    & Bottom Up    & 1000 & 26.66\% &\textbf{34.99\%}& 30.35\% &        60.11\% &     78.14\% & 67.61\% &        72.85\% &     92.94\% & 81.30\% \\ \midrule
\multirow{3}{*}{\textbf{Least Squared Deviation}}
    & Window Based & 1000 & 25.01\% &    16.14\% & 20.26\% & 54.58\% &     35.77\% & 42.96\% & 68.73\% &     48.11\% & 56.18\% \\
    & Pelt         & 1000 & 21.28\% &    80.04\% & 33.49\% & 51.05\% &     99.58\% & 67.20\% & 65.36\% & \textbf{99.97\%} & 78.76\% \\
    & Bottom Up    & 1000 & 21.30\% &    81.24\% & \textbf{33.62\%}  &     50.98\% &\textbf{99.50\%}& 67.12\% &        65.24\% &\textbf{99.97\%}& 78.66\% \\ \midrule
\multirow{3}{*}{\textbf{Linear Model Change}}
    & Bottom Up    & 100  &  7.39\% &     0.39\% &  9.98\% & 27.44\% &      1.49\% & 10.27\% & 52.51\% &      2.75\% &  9.90\% \\
    & Window Based & 100  &  7.39\% &     0.39\% &  9.98\% & 27.44\% &      1.49\% & 10.27\% & 52.51\% &      2.75\% &  9.90\% \\
    & Pelt         & 100  &  7.39\% &     0.39\% &  9.98\% & 27.44\% &      1.49\% & 10.27\% & 52.51\% &      2.75\% &  9.90\% \\ \midrule
\multirow{3}{*}{\textbf{Gaussian Process Change}}
    & Window Based & 100  &  7.39\% &     0.39\% &  9.98\% & 27.44\% &      1.49\% & 10.27\% & 52.51\% &      2.75\% &  9.90\% \\
    & Pelt         & 100  &  7.39\% &     0.39\% &  9.98\% & 27.44\% &      1.49\% & 10.27\% & 52.51\% &      2.75\% &  9.90\% \\
    & Bottom Up    & 100  & 12.42\% &\textbf{100.00\%}& 22.03\% &33.15\% &\textbf{100.00\%}& 49.55\% &        49.04\% &    \textbf{100.00\%} & 65.47\% \\ \midrule
\multirow{2}{*}{\textbf{Rank-based Cost Function}}
    & Pelt         & 100  & 21.11\% &    31.85\% & 25.51\% & 52.29\% &     74.63\% & 61.13\% & 65.88\% &     89.30\% & 75.46\% \\
    & Bottom Up    & 100  & 20.55\% &    31.84\% & 25.17\% & 49.97\% &     73.22\% & 59.04\% & 64.57\% &     89.35\% & 74.61\% \\ \midrule
\multirow{3}{*}{\textbf{Kernelized Mean Change}}
    & Bottom Up    & 100  & 15.66\% &     1.90\% &  9.45\% & 44.23\% &      5.36\% & 12.95\% & 62.37\% &      7.48\% & 15.66\% \\
    & Pelt         & 100  & 14.07\% &     1.64\% &  9.77\% & 42.06\% &      4.74\% & 12.59\% & 60.92\% &      6.67\% & 14.56\% \\
    & Window Based & 100  &  8.31\% &     0.61\% &  9.97\% & 29.38\% &      2.02\% & 10.60\% & 53.34\% &      3.39\% & 10.79\% \\ \midrule \midrule
\multirow{3}{*}{\textbf{Our Approach}}
    & \multirow{3}{*}{\textbf{:}}
         & \textbf{Test}    &  \textbf{44.70\%} &    52.38\% & \textbf{48.24\%} &   \textbf{84.56\%} &     90.54\% & \textbf{87.45\%} &   \textbf{90.53\%} &     96.86\% & \textbf{93.59\%} \\
    & {} & \textbf{Validation (Tuning)} &  42.56\% &    48.26\% & 45.23\% &   87.11\% &     89.68\% & 88.37\% &   92.54\% &     96.48\% & 94.47\% \\
    & {} & \textbf{Training}      &  56.43\% &    62.53\% & 59.32\% &   89.64\% &     92.79\% & 91.19\% &   93.97\% &     97.93\% & 95.91\% \\
\bottomrule
\end{tabular}%
}
\end{table*}

\subsubsection{RQ2.3 Results: Multi-class Classification Performance on Paparazzi (replication)}
Table~\ref{tab:rq2-1-results-replicate} presents the comparative results of the baseline algorithms and those of our approach.
In comparison with the MP case, we can see that in all but two settings limiting the maximum number of features did not help improve the model. A further observation is that the scores do not vary much from changing the window size, and decision trees almost universally outperform linear classifiers.
The scores themselves are lower: The Ridge classifier F1 score here is in 21-29\% range, which was 32-39\% in RQ1.2. For decision trees it is in 67-68\% range here while it is in 73-77\% range in RQ1.2.

The deep learning model's performance measures (on test data) can be found in the last row of the table.
There is an $(82.23/68.96)-1=19.24\%$ improvement in F1 score over the baselines. 
The improvement in RQ1.2 is 16.83\%, not very different from this. This, again provides some evidence on the generalizability of the results in the first two RQs.

\begin{table}[t]
\caption{Precision, recall, and F1 score of ridge classifiers (linear classifiers with L2 regularization) and decision tree classifiers with different sliding window widths ($w$), on Paparazzi dataset. In this table, we only show the results of the best performing model in each group.}
\label{tab:rq2-1-results-replicate}
\resizebox{\linewidth}{!}{%
\begin{tabular}{llcccc}
\toprule
\textbf{w} &
  \multicolumn{1}{c}{\textbf{Classifier}} &
  \multicolumn{1}{c}{\textbf{Max Features}} &
  \multicolumn{1}{c}{\textbf{Precision}} &
  \multicolumn{1}{c}{\textbf{Recall}} &
  \multicolumn{1}{c}{\textbf{F1}} \\ \midrule
 3 &  Ridge &            - &      44.82\% &   14.45\% & 21.85\% \\
 3 & Decision Tree & $\sqrt{10w}$ &      61.88\% &   73.56\% & 67.22\% \\ \midrule
 5 &  Ridge &            - &      46.62\% &   15.10\% & 22.81\% \\
 5 & Decision Tree &            - &      64.28\% &   73.00\% & 68.36\% \\ \midrule
10 &  Ridge &            - &      47.59\% &   16.29\% & 24.27\% \\
10 & Decision Tree &            - &\textbf{65.06\%}& 73.36\% &\textbf{68.96\%}\\ \midrule
15 &  Ridge &            - &      44.41\% &   17.33\% & 24.93\% \\
15 & Decision Tree &            - &      63.53\% &   74.68\%& 68.65\% \\ \midrule
20 &  Ridge &            - &      57.97\% &   19.10\% & 28.74\% \\
20 & Decision Tree &            - &      64.72\% &   73.36\% & 68.77\% \\ \midrule
25 &  Ridge &            - &      62.13\% &   19.66\% & 29.87\% \\
25 & Decision Tree & $\sqrt{10w}$ &      61.17\% &\textbf{75.18}\% & 67.45\% \\ \midrule \midrule
   & \multicolumn{2}{l}{Proposed Method} & \textbf{80.86\%} & \textbf{88.23\%} & \textbf{84.38\%} \\
\bottomrule
\end{tabular}%
}
\end{table}

\begin{figure*}
    \centering
    \includegraphics[width=\textwidth]{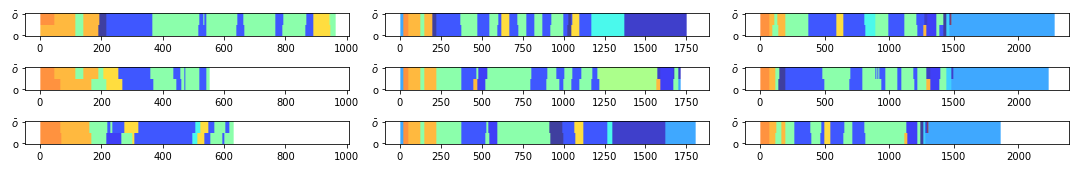}
    \caption{The states are color coded and each rectangle shows one test, each column contains 3 samples from different flight lengths consisting of 3 groups, short($|l|$$<$1400; the first tertile), intermediate(1400\textless$|l|$\textless1800; the second tertile) and long(1850\textless $|l|$). The same as figure~\ref{fig:test_0} in, the top half of each rectangle shows the model's output and the bottom half shows the true labels.}
    \label{fig:paparazzi_predictions}
\end{figure*}

Figure~\ref{fig:paparazzi_predictions} summarizes the prediction results of the deep learning model compared to the ground truth.

\begin{rqanswer}
Compared to the baselines, the proposed approach could detect the internal state of system with a 77\% precision and 88\% recall rate, showing a 19\% improvement on baselines. Similar results were seen in RQ 1.2 (which is closely related to this question) which provide evidence on the generalizability of the findings, to other UAV autopilot systems.
\end{rqanswer}

\subsubsection{RQ2.4 - Transfer Learning}
\begin{table}[]
\caption{Results of the model trained using transfer learning (only dense layers) on 5 Paparazzi data points (Transfer Learning) vs. all layers trained on 5 Paparazzi data points, i.e., no transfer learning (Training a full model)}
\label{tab:tl_all_meansures}
    \centering
\resizebox{\linewidth}{!}{%
    \begin{tabular}{lcc}
    \toprule
\textbf{Evaluation Metric} & \textbf{Transfer Learning} & \textbf{Training a full model} \\\midrule
Precision ($\tau=1s$)      &  24.76\%  &  21.42\% \\
Recall ($\tau=1s$)         &  29.20\%  &  22.50\% \\
F1 ($\tau=1s$)             &  26.73\%  &  20.99\% \\ \midrule
Precision ($\tau=3s$)      &  56.44\%  &  49.63\% \\
Recall ($\tau=3s$)         &  60.16\%  &  46.50\% \\
F1 ($\tau=3s$)             &  58.12\%  &  45.85\% \\ \midrule
Precision ($\tau=5s$)      &  74.78\%  &  69.87\% \\
Recall ($\tau=5s$)         &  74.14\%  &  59.96\% \\
F1 ($\tau=5s$)             &  74.35\%  &  61.62\% \\ \midrule
State Detection Prec.  &  73.12\%  &  72.65\% \\
State Detection Recall     &  75.82\%  &  70.20\% \\
State Detection F1         &  74.40\%  &  71.23\% \\
\bottomrule
    \end{tabular}%
}
\end{table}

As discussed, to evaluate the generalizability of our approach on cases where there are not many labeled data available, we study transfer learning (TL). Table~\ref{tab:tl_all_meansures} summarizes TL results by comparing the 12 evaluation metrics we have been using in previous RQs, when TL is employed, compared to training a model from scratch (called ``Training a full model''). 
The results shows that the transfer learning method improves the baseline (``Training a full model'') for all reported metrics. The improvements over the baseline are in the range of 1\% to 30\%. For instance looking at the F1 scores, using TL (with only 5 labeled test cases) we can achieve 74.35\% for CPD with $\tau=5s$ and 74.40\% for state classification tasks.    

To put these numbers into perspective, the maximum F1 scores of our model with a fully labeled training set is 93.59\% and 82.23\%, for the CPD (with $\tau=5s$) and the classification tasks, respectively (See Tables ~\ref{tab:cpd_paparazzi} and ~\ref{tab:rq2-1-results-replicate}). That means TL can achieve 79\%, and 90\% of the maximum potentials, for the same tasks, by much less manual effort (98\% labeling cost reduction).

\begin{rqanswer}
Using transfer learning up to 90\% of the maximum potentials of our model (in terms of F1 scores) can be be achieved with only 2\% of manual labeling costs. 

\end{rqanswer}

\subsection{Limitations and Threats to Validity} \label{sec:threats_to_validity}

\subsubsection{Limitations of the Approach}

One of the limitations of this approach is that it might miss an input-output invariant correlation. It can happen when the input remains constant or it changes too little to reveal its relation with certain outputs.

We assume that during the data collection, sampling happens in regular intervals; this approach probably will have a hard time achieving high performances, working on unevenly spaced time-series data.

Another limitation is that this approach requires the entire dataset at once to be able to infer the models. In other words, the model can not handle a stream of data as they arrive, thus limiting its application for use cases such as anomaly detection at run-time. We intend to address this in future work by adapting the model architecture to an online learning model.

\subsubsection{Threats to Validity}

In terms of construct validity, we are using standard metrics to evaluate the results. However, the use of tolerance margin should be treated with caution since it is a domain-dependent variable and can change the final results. To alleviate this threat, we have used multiple margins and reported all results.

Another potential construct validity threat is the fact that we used the simulator to provide an authoritative labelling of our data, whereas in practice this would be carried out by a domain expert in a fully black-box manner. This is a realistic expectation, since monitoring the logs and identifying the current system state is part of the developers/testers regular practice during inspection and debugging. 
In addition, the transfer learning approach reduces the need for manual labeling significantly, when the approach is being reused on similar projects (e.g., different products in a product line). Further study of the amount of mislabeling in practice and their effect on the final results and the potential time needed for detection and adjustment requires a separate user study, which is beyond the scope of this paper.

Also note that, even though we use the source code to label the training set, we still treat the data as though it were obtained from a black-box and do not take advantage of the additional information that one could obtain from source code.

In terms of threats to internal validity, we reduced the potential for bias against baseline approaches by not implementing the CPD baselines ourselves and reusing existing libraries.

In terms of conclusion validity, one threat is that we draw incorrect conclusions from  limited observations. To address this risk we based our results on a  large test suite of 888 real test cases from MicroPilot's test repository, and provided a proper train-validation-test split for training, tuning, and evaluation.

Finally, in terms of external validity threats, this study is limited to only two case studies. However, (a) the studies are large-scale real-world systems with many test cases, and (b) they are both from industry and open source to be more representative. Our future work will look to extend this work with more case studies from other domains, to increase its generalizability.

\section{Conclusion and Future Work}\label{sec:summary} \label{sec:future_work}
In this paper, we developed a novel method for inferring black box models for autopilot software systems.
Our method at its core is a deep neural network that combines convolutions and recurrent cells: hybrid CNN-RNN model. 
This design is inspired by deep neural network architectures that showed good performance in other fields (such as speech recognition, sleep phase detection, and human activity recognition). 
It can be used for both CPD and state classification problems in multivariate time series. 
This method can be used as a black-box state model inference for variety of use cases such as testing, debugging, and anomaly detection in control software systems, where there are several input signals that control output states. 
We have trained and evaluated this neural network on two case studies of UAV autopilot software, one open-Source and one from our industry partner.
It showed promising results in inferring a behavioural model from the autopilot execution data; showing significant improvement in both change point detection and state classification compared to several baselines on 10 comparison metrics. Our proposed transfer learning approach also help reducing the manual labeling cost, significantly.
Some potential extensions to this work include:
(a) adapting this method to other domains such as self-driving cars, (b) using the inferred models to perform a downstream task such as anomaly detection, and (c) improving the hyper-parameter optimization part with more advanced tuning techniques.

\ifCLASSOPTIONcompsoc
  \section*{Acknowledgments}
\else
  \section*{Acknowledgment}
\fi

We acknowledge the support of the Natural Sciences and Engineering Research Council of Canada (NSERC), [funding reference number CRDPJ/515254-2017]. Walkinshaw is supported by the Engineering and Physical Sciences Research Council in the UK (EPSRC), [CITCOM - EP/T030526/1].

\ifCLASSOPTIONcaptionsoff
  \newpage
\fi



%
\bibliographystyle{IEEEtran}
\bibliography{bibliography}

%








\end{document}